\documentclass{article}
\usepackage{arxiv}
\usepackage[utf8]{inputenc} % allow utf-8 input
\usepackage[T1]{fontenc}    % use 8-bit T1 fonts
\usepackage{hyperref}       % hyperlinks
\usepackage{url}            % simple URL typesetting
\usepackage{booktabs}       % professional-quality tables
\usepackage{nicefrac}       % compact symbols for 1/2, etc.
\usepackage{microtype}      % microtypography
\usepackage{lipsum}         % Can be removed after putting your text content
\usepackage{graphicx}
\usepackage{natbib}
\usepackage{doi}

\usepackage{researchpack}
\usepackage[capitalize,noabbrev,nameinlink]{cleveref}
\usepackage[dvipsnames]{xcolor}
\usepackage{enumitem}
\usepackage{xspace}
\usepackage{multirow}
\usepackage{systeme}
\usepackage{arydshln}

\usepackage{tikz}
\usetikzlibrary{positioning,shapes,arrows,calc}

\definecolor{ClevrGray}{RGB}{128,128,128}
\definecolor{ClevrGreen}{RGB}{35,127,21}
\definecolor{ViolinBlue}{RGB}{0, 123, 167}
\definecolor{ViolinRed}{RGB}{255, 67, 164}
\definecolor{NavyColor}{RGB}{0, 0, 128}

\hypersetup{
    colorlinks=true,
    citecolor=NavyColor,
    linkcolor=NavyColor,
    filecolor=NavyColor,
    urlcolor=NavyColor,
    pdftitle={Neuro-Symbolic Reasoning Shortcuts: Mitigation Strategies and their Limitations},
    pdfauthor={Emanuele Marconato, Stefano Teso, Andrea Passerini},
    pdfkeywords={Machine Learning, Neuro-Symbolic Integration, Structured-output Prediction, Concepts, Reasoning Shortcuts},
}

\newcommand{\MZero}{\includegraphics[width=1.85ex]{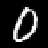}\xspace}
\newcommand{\MOne}{\includegraphics[width=1.85ex]{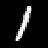}\xspace}
\newcommand{\MTwo}{\includegraphics[width=1.85ex]{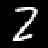}\xspace}

\newcommand{\dataset}{\ensuremath{\calD}\xspace}

\newcommand{\BK}{\ensuremath{\mathsf{K}}\xspace}

\title{Neuro-Symbolic Reasoning Shortcuts:\\ Mitigation Strategies and their Limitations}

\date{March 21, 2023}

\author{
    Emanuele Marconato\thanks{Corresponding author.} \\
    DISI, University of Trento, Italy \\
    University of Pisa, Italy \\
    \texttt{emanuele.marconato@unitn.it} \\
    \And
    Stefano Teso \\
    CIMeC and DISI, University of Trento, Italy \\
    \texttt{stefano.teso@unitn.it} \\
    \And
    Andrea Passerini \\
    DISI, University of Trento, Italy \\
    \texttt{andrea.passerini@unitn.it} \\
    }

\renewcommand{\shorttitle}{Neuro-Symbolic Reasoning Shortcuts}

\begin{document}

\maketitle
\begin{abstract}
    Neuro-symbolic predictors learn a mapping from sub-symbolic inputs to higher-level concepts and then carry out (probabilistic) logical inference on this intermediate representation.
    This setup offers clear advantages in terms of consistency to symbolic prior knowledge, and is often believed to provide interpretability benefits in that -- by virtue of complying with the knowledge -- the learned concepts can be better understood by human stakeholders.
    However, it was recently shown that this setup is affected by \textit{reasoning shortcuts} whereby predictions attain high accuracy by leveraging concepts with unintended semantics~\citep{marconato2023neuro,li2023learning}, yielding poor out-of-distribution performance and compromising interpretability.
    In this short paper, we establish a formal link between reasoning shortcuts and the optima of the loss function, and identify situations in which reasoning shortcuts can arise.
    Based on this, we discuss limitations of natural mitigation strategies such as reconstruction and concept supervision.
\end{abstract}



\section{Introduction}

Neuro-symbolic (NeSy) integration of learning and reasoning is a key challenge in AI.
NeSy \textit{predictors} achieve integration by learning a \textit{neural network} mapping low-level representations (e.g., MNIST images) to high-level symbolic concepts (e.g., digits), and then predicting a label (e.g., the sum) by \textit{reasoning} over concepts and prior knowledge~\citep{manhaeve2018deepproblog}.
Most works on the topic focus on how to best integrate knowledge into the loop, cf.~\citep{de2021statistical}.  The issue of \textit{concept quality} is, however, generally neglected.  Loosely speaking, the consensus is that knowledge ensures learning high quality concepts and that issues with these should be viewed as ``learning artifacts''.

This is not the case.  Recently, \citet{li2023learning} and \citet{marconato2023neuro} have shown that NeSy predictors can learn \textit{reasoning shortcuts} (RSs), that is, mappings from inputs to concepts that yield high accuracy on the training set by predicting the \textit{wrong} concepts.
While RSs -- by definition -- do not hinder the model's accuracy on the training task, they prevent identification of concepts with the ``right'' semantics, and as such compromise \textit{generalization} beyond the training distribution and \textit{interpretability}~\citep{marconato2023neuro}.
As an example, consider MNIST Addition~\citep{manhaeve2018deepproblog}.  Here, the model has to determine the sum of two MNIST digits, under the constraint that the sum is correct.
Given the examples ``$\MZero + \MOne = 1$'' and ``$\MZero + \MTwo = 2$'', there exist two alternative solutions:  the intended one $(\MZero \to 0, \MOne \to 1, \MTwo \to 2)$ and a RS $(\MZero \to 1$, $\MOne \to 0$, $\MTwo \to 1)$. 
Both of them ensure the sum is correct, but only one of them captures the correct semantics.

This begs the question:  \textit{under what conditions do reasoning shortcuts appear, and what strategies can be used to mitigate them?}
In this short paper, we outline answers to these questions.
First, we go beyond existing works and show how to \textit{count} the number of RSs affecting a NeSy prediction task.
Based on this result, we show that, in the general case, it is \textit{impossible} to identify the correct concepts from label supervision only.
We also consider two mitigation strategies, namely reconstruction and concept supervision, and study their effects and limitations.

\section{Neuro-symbolic task construction}

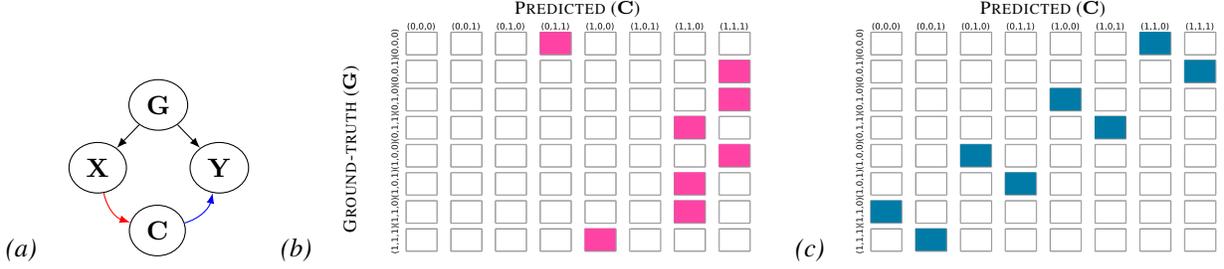
\begin{figure}[!t]
    \centering          
    \begin{tabular}{rcrccrc}
        % \textit{(a)} & \textit{(b)} & \textit{(c)}  \\
        & & & & {\scriptsize \textsc{Predicted} ($\vC$)} & & {\scriptsize \textsc{Predicted} ($\vC$)} \\
        \textit{(a)} &
        \begin{tikzpicture}[
            scale=1,
            transform shape,
            node distance=.35cm and .25cm,
            minimum width={width("Xtil")+2pt},
            minimum height={width("Xtil")+2pt},
            mynode/.style={draw,ellipse,align=center}
        ]
            \node[mynode] (Gi) {$\vG$};
            \node[mynode, below left=of Gi] (X) {$\vX$};
            \node[mynode, below right=of Gi] (Y) {$\vY$};
            \node[mynode, below right=of X] (C) {$\vC$};

            \path (Gi) edge[-latex] (X)
            (Gi) edge[-latex] (Y);
            % (X) edge[-latex] (C);
            \path [red, bend right](X) edge[-latex] (C);
            \path [blue,bend right](C) edge[-latex] (Y);
            % \path [red,bend right] (Gi) edge[-latex] (C);
        \end{tikzpicture} & \textit{(b)} & \rotatebox{90}{\hspace{0.65em} \scriptsize \textsc{Ground-truth} ($\vG$) } & 
        \includegraphics[width=0.3\textwidth]{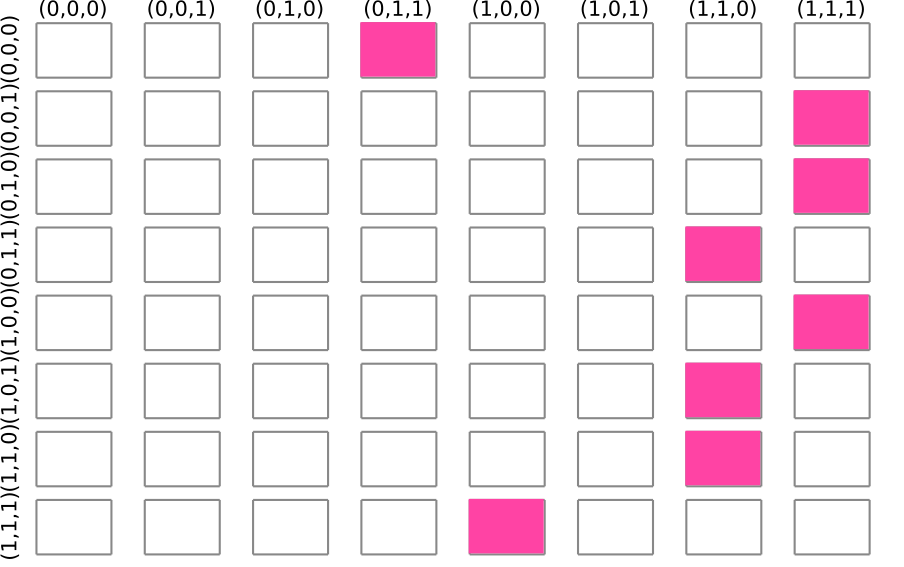} & \textit{(c)} &
        \includegraphics[width=0.3\textwidth]{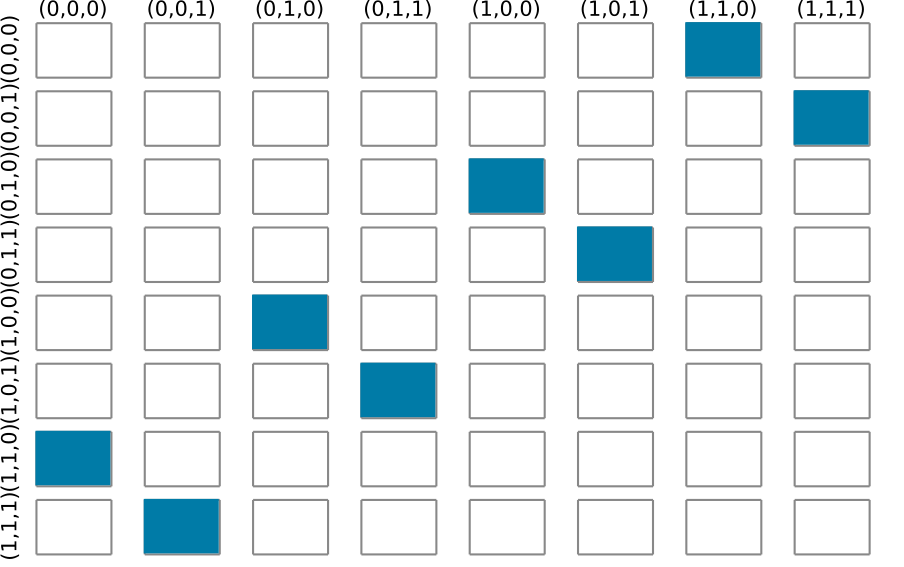} 
    \end{tabular}

    \caption{
    \textit{(a)} Graphical model of our setup:  the \textbf{black} arrows encode the data generation process, the \textcolor{red}{\textbf{red}} arrow indicate the learned concept distribution, and the \textcolor{blue}{\textbf{blue}} arrow the reasoning module.
    \textit{(b)} Confusion matrices of (Boolean) concepts learned by DeepProblog for our three bits XOR task without any mitigation strategy, and
    \textit{(c)} with a reconstruction term, cf. \cref{eq:solutions-reasoning-rec}.
    In both cases the learned concepts are consistent with the knowledge and in the second one they also manage to reconstruct the input.  %Additionally, reconstruction allows to map distinct ground-truth concept vectors into distinct predicted concept vectors.
    The confusion matrices immediately show that, despite this, the learned concepts are reasoning shortcuts.
    }
    \label{fig:generative-process}
\end{figure}

We consider a NeSy prediction task where, given sub-symbolic inputs $\vX$, the goal is to infer one or more labels $\vY \in \{0, 1\}^\ell$ consistent with a given propositional formula $\BK$ encoding prior knowledge.
We focus on DeepProbLog~\citep{manhaeve2018deepproblog}, a representative and sound framework for such tasks.
From a probabilistic perspective, DeepProbLog:
(i) Extracts $k$ concepts $\vC \in \{0, 1\}^k$ from a $\vX$ via a neural network $p_\theta (\vC \mid \vX)$, and
(ii) Models the distribution over the labels $\vY$ as a uniform $u_\BK (\vy \mid \vc) = \Ind{(\vy, \vc) \models \BK}$.
The label distribution is obtained by marginalizing $\vC$:
\[
    \textstyle
    p_\theta(\vy \mid \vx; \BK) = \sum_\vc u_\BK(\vy \mid \vc) p_\theta(\vc \mid \vx)
    \label{eq:deepproblog}
\]
DeepProbLog is then trained via \textit{maximum likelihood}. \hspace{0.3em}
In order to understand when doing so recovers concepts $\vC$ with the ``correct semantics'', we have to first define the unobserved generative mechanism underlying the training data whose concepts we wish to identify.
Motivated by work on identifiability in (causal) representation learning \citep{locatello2019challenging, scholkopf2021toward, khemakhem2020variational, ahuja2022towards}, we assume there exist $k$ ground-truth concepts $\vG \in \{0, 1\}^k$ spanning a space $\calG$,
%\footnote{Notice that the support of $\vC$ corresponds to all concepts accounted in the knowledge $\BK$, so it coincides with $\calG$. \ST{unclear; is this necessary?}} 
and that the examples $(\vX, \vY) = (f( \vG ), h(\vG))$ are generated by an invertible function $f: \calG \to \mathcal X \subset \bbR^d$ and a surjective function $h: \calG \to \calY$, with $|\calY| \leq |\calG|$.
Here, $h$ plays the role of the ground-truth reasoning module that infers the label $\vY$ from the ground-truth concepts $\vG$ according to $\BK$, while $f$ generates the observations themselves.\footnote{Due to space constraints, we assume $\vX$ depends on $\vG$ only.  In practice, it might also depend on additional ``stylistic'' factors of variation (e.g., font)~\citep{von2021self}. Our results apply to this more complex case with minimal modifications.}
Cf. \cref{fig:generative-process} for an illustration.
In the next sections, we will show how maximum likelihood training can recover the mechanism $g \circ f^{-1}$, but not the ground-truth mapping from inputs to concepts $f^{-1}$, \ie the ``correct semantics''.

% Concerning the inference mechanism, we consider a factorized distribution $p_\theta (\vC, \vY \mid \vX; \BK) =  p_\theta(\vY \mid \vC; \BK)  p_\theta(\vC|\vX)$ with learnable parameters $\theta$. 
% %
% Here, the variables $\vC$ represent the learned concepts which are typically encoded via neural networks. Without loss of generality, the prediction of $\vY$ depends solely on the concepts $\vC$, since they contain all relevant information for the classification problem. 
% %
% For instance, DeepProblog \citep{manhaeve2018deepproblog} infers $\vY$ by grounding a probabilistic logic program based on the knowledge $\BK$ on the ...

% In the rest of the paper, we consider only the probability decomposition given by DeepProblog, and we leave a general characterization for future works.

\section{Reasoning shortcuts and mitigation strategies}

We consider training points $(\vx, \vy) \in \dataset_{\vX, \vY}$, each originated by corresponding ground-truth concepts $\vg \in \dataset_\vG$.\footnote{We assume the training examples are \textit{noiseless} and cover all possible combinations of ground-truth factors $\vG$, as even this ``ideal'' setting admits RSs.}  Our starting point is the log-likelihood, which constitutes the objective of training:
\[
    \textstyle
    \calL(\theta) := \sum_{ (\vx,\vy) \in \dataset_{\vX, \vY}} \log p_\theta(\vy \mid \vx; \BK) \equiv \sum_{ \vg \in \dataset_{\vG} } \log  p_\theta \big( h(\vg) \mid f(\vg); \BK  \big)
    \label{eq:log-lh-nesy}
\]
Notice that all optima of \cref{eq:log-lh-nesy} satisfy $p_\theta (\vy \mid \vx; \BK  ) = 1$ for all examples.  By \cref{eq:deepproblog}, this entails that any $\vc \sim p_\theta( \vc \mid \vx )$ must satisfy the knowledge $\BK$, that is, $(\vc, \vy) \models \BK$ (see \citep[Theorem 3.2]{marconato2023neuro}).
How many alternative distributions $p_\theta(\vc \mid \vg) := p_\theta(\vc \mid f(\vg))$ do attain maximum likelihood?  Since $p_\theta$ is a neural network, there may be infinitely many, yet all of them except one are RSs.  This is sufficient to show that \textit{RSs cannot be discriminated from the ground-truth concept distribution based on likelihood alone}~\citep{marconato2023neuro}.

Importantly, it turns out all optimal distributions $p_\theta(\vc \mid \vx)$ are convex combinations of the \textit{deterministic} optima (\textit{det-opts}), that is, those distributions $p_\theta(\vc \mid \vg)$ mapping each $\vg$ to a unique $\vc$ with probability one.
If the likelihood admits a single \textit{det-opt}, this is also the \textit{only} solution and -- by construction -- it recovers the ground-truth concepts.  \textit{RSs arise when there are two or more det-opts}.
How many det-ops are there?  Let $S_\vy = \big \{ \vc \, : \, (\vc, \vy) \models \BK \big \}$ be the set of $\vc$'s that $\BK$ assigns to label $\vy$.
Notice that if $p_\theta(\vc \mid \vg)$ attains maximum likelihood, then any $\vc \sim p_\theta(\vc \mid \vg)$ falls within $S_{h(\vg)}$.
In this sense, a \textit{det-opt} implicitly maps each vector $\vg \in \calG$ to a vector $\vc \in S_{h(\vg)}$.
This gives us a mechanism to count \textit{det-opts}: for each $\vg$ there are exactly $|S_{h(\vg)}|$ vectors $\vc$ that it can be mapped to, meaning that number of \textit{det-opts} for \cref{eq:log-lh-nesy} is:
\begin{equation}
    \textstyle
    \#\text{det-opts}(\calL) = \prod_{\vy \in \calY} |S_\vy |^{|S_\vy|}
    \label{eq:solutions-to-reasoning}
\end{equation}
As a consequence, the ground-truth concepts can only be retrieved if $|S_\vy| = 1$, \ie each label $\vy$ can be deduced from a unique $\vc$.
%(and, consequently, $|\calY| = |\calG|$).
%
This is seldom the case in NeSy tasks, meaning that maximizing the likelihood of the labels $\vY$ cannot rule out RSs in general.

In the following, we discuss two natural mitigation strategies and their impact in reducing the total number of \textit{det-ops}.

\paragraph{Reconstruction is insufficient.}  Given the likelihood is incapable of discriminating intended and RS solutions, one option is to augment it with a term encouraging learned concepts $\vC$ to capture information necessary to reconstruct the input $\vX$, for instance:
\[
    \textstyle
    \calR (\theta) := \sum_{\vx \in \dataset_\vX} \big[ \sum_{\vc} p_\theta (\vc \mid  \vx) \log p_\psi( \vx \mid \vc ) \big] \equiv \sum_{\vg \in \dataset_\vG} \big[ \sum_{\vc} p_\theta (\vc \mid  \vg) \log p_\psi( \vg \mid \vc )  \big]
    \label{eq:rec-term}
\]
Here, $p_\psi(\vx \mid \vc)$ is the distribution output by a neural decoder with parameters $\psi$, and we introduced $p_\psi (\vg \mid \vc) := p_\psi(f(\vg) \mid \vc)$. 
The optima of \cref{eq:rec-term} must satisfy $p_\psi(\vg \mid \vc) = 1$ for all $\vc \sim p_\theta(\vc \mid \vg)$.
In other words, restricting again to \textit{det-ops} for the encoder, the only \textit{det-ops} that ensure perfect reconstruction are those mapping distinct $\vg$'s to distinct $\vc$'s, \ie that ensure the encoder is injective.
How many such \textit{det-opts} are there?  Notice that these \textit{det-opts} can be enumerated by taking each $\vg \in \calG$ in turn and mapping it to an arbitrary $\vc$ in $S_{h(\vg)}$ \textit{without replacement} (to ensure injectivity), until all $\vg$'s have been mapped.
%
%When combining the two objectives in \cref{eq:log-lh-nesy} and \cref{eq:rec-term}, a \textit{det-opt} implicitly defines a permutation on the elements of $S_{\vy}$, so that the number of possible solutions amounts to $|S_\vy |!$ for each $\vy$.
%
This entails that the number of \textit{det-opt}s -- under perfect reconstruction -- becomes:
\[
    \textstyle
    \#\text{det-opts}(\calL + \calR) = \prod_{\vy \in \calY} |S_\vy| !
    \label{eq:solutions-reasoning-rec}
\]
%
% Notice that, in this case, convex combinations of \textit{det-opts} are no longer valid solution as they would not optimize the reconstruction term.
Once again, unless $|S_\vy|=1$ for all $\vy$'s, there are multiple possible solutions, most of which are RSs.
% Nonetheless, the scaling of the possible solutions remains factorial with the dimension of the equivalence classes, so
In other words, \textit{adding a reconstruction term can be insufficient to completely rule out learning reasoning shortcuts}.

\paragraph{The effect of concept supervision.}  Next, we consider a scenario where \textit{concept supervision} is provided (for all concepts) for at least some examples $(\vx, \vg) \in \overline{\dataset}_{\vX, \vG}$.  We consider the $L_2$ loss for fitting the supervision, for simplicity:
\[
    \textstyle
    \calC (\theta) \propto  \sum_{(\vx, \vg) \in \overline{\dataset}_{\vX, \vG} }  ( \vc - \vg)^2 p_\theta (\vc \mid \vx)  \equiv \sum_{\vg \in \overline{\dataset}_{\vG}  }\sum_{\vc} ( \vc - \vg)^2 p_\theta (\vc \mid \vg) 
    \label{eq:concept-supervision-loss}
\]
The only concept distributions $p_\theta(\vc \mid \vg)$ minimizing \cref{eq:concept-supervision-loss} are those that allocate all probability mass to the annotated concepts.
Now, let $\nu_\vy$ be the number of vectors $\vc \in S_\vy$ for which we have supervision $\vg$, for a total of $|\overline{\dataset}_{\vG}| = \sum_\vy \nu_\vy$.
The situation is analogous to \cref{eq:solutions-to-reasoning} and \cref{eq:solutions-reasoning-rec}, except that now for exactly $\nu_\vy$ vectors $\vc$ we know exactly what $\vg$ they should be mapped to, leaving the remaining $|S_\vy| - \nu_\vy$ vectors dangling.  This gives:
% Upon optimizing \cref{eq:concept-supervision-loss} we disambiguate the concepts in $\overline{\dataset}_\vG$, Hence, the number of total \textit{det-opt}s becomes:
%
\[
    \textstyle
     \# \text{det-opts}(\calL + \calC)  = \prod_{\vy \in \calY} |S_\vy|^{ |S_\vy| - \nu_\vy  }, \quad \quad \# \text{det-opts}( \calL + \calR + \calC)  = \prod_{\vy \in \calY} ( |S_\vy| - \nu_\vy )!
     \label{eq:total-sol-concepts}
\]
Here, the first term counts how many \textit{det-opt}s optimize both the label likelihood and the concept supervision, and the second one those optimizing the likelihood, reconstruction and concept supervision.
This shows providing concept supervision can dramatically reduce the number of \textit{det-opts} but also that \textit{a substantial amount is necessary to rule out all RSs}. \newpage
%
% When combining all terms, the number of total concepts that need to be supervised to identify the ground-truth mechanism $f^{-1}$ amounts to: $ \nu = \sum_{\vy \in \calY} (|S_\vy| -1) = |\calG \mid - |\calY| $.

\section{Empirical Verification}

We outline a toy experiment showing how reasoning shortcuts affect even a simple NeSy task. 
Let $\vg = (g_1, g_2, g_3)$ be three bits and consider the task of predicting their parity, that is, $y = g_1 \oplus g_2 \oplus g_3$.
Each label $y \in \{0, 1\}$ can be deduced from $4$ possible concept vectors $\vg$.
We train two MLPs, one encoding directly $\vg$ into $p_\theta (\vc \mid \vg)$, and another decoding $\vc$ into $p_\psi(\vg \mid \vc)$.  Labels are predicted as per \cref{eq:deepproblog}.
Given the problem at hand, the total number of \textit{det-opt}s is given by \cref{eq:solutions-to-reasoning} and by \cref{eq:solutions-reasoning-rec}:
$$ \# \text{det-opts}(\calL) = (4^4 \cdot 4^4) \quad \quad
\# \text{det-opts}(\calL + \calR) = (4! \cdot 4!)
$$
Empirically, what happens is that without concept supervision, the model picks up reasoning shortcuts to solve the task.
\cref{fig:generative-process} shows two such RSs, both optimal, obtained by our model when optimizing \textit{(b)} only the likelihood, and \textit{(c)} both the likelihood and the reconstruction term.  In both cases, the solutions fail to recover the ground-truth concepts.

\section*{Conclusion.} Our results altogether show that the ground-truth concepts are hard, if not impossible, to recover empirically, and that two natural mitigation strategies do not completely address the problem.
In particular, the amount of concept supervision required grows linearly with the number of possible concept combinations.
We envisage well-tuned strategies based on targeted concept-supervision, combined with additional restrictions on the model itself (and specifically \textit{disentanglement} between concepts~\citep{suter2019robustly}), will likely facilitate (provable) identification of the ground-truth concepts.
This is left to future work.

\section*{Acknowledgements}

The research of ST and AP was partially supported by TAILOR, a project funded by EU Horizon 2020 research and innovation programme under GA No 952215.

%%
%% Define the bibliography file to be used
\bibliographystyle{unsrtnat}
\bibliography{references.bib}


\end{document}